\pdfoutput=1

\documentclass[10pt,twocolumn,letterpaper]{article}

\usepackage{iccv}
\usepackage[utf8]{inputenc}
\usepackage{amsmath} 
\usepackage{makecell} 
\usepackage{mathptmx} 
\usepackage{booktabs} 
\usepackage{caption}
\usepackage{subcaption}
\usepackage{graphicx} 
\usepackage{amssymb} 
\usepackage{multirow} 

\usepackage{color}
\definecolor{citecolor}{HTML}{0071bc}
\usepackage{colortbl}  

\usepackage[pagebackref=true,breaklinks=true,letterpaper=true,citecolor=citecolor,colorlinks,bookmarks=false]{hyperref}
\newcommand{\tablestyle}[2]{\setlength{\tabcolsep}{#1}\renewcommand{\arraystretch}{#2}\centering\small}
\newcommand{\myparagraph}[1]{{\noindent\bf #1}}
\newcommand{\methodname}{P3Former}
\newcommand{\eqnsm}[2]{\begin{equation}\label{eq:#1}#2\end{equation}}
\newcommand{\TP}{\mathit{TP}}
\newcommand{\FN}{\mathit{FN}}
\newcommand{\FP}{\mathit{FP}}

\usepackage{threeparttable}
\usepackage{caption} 

\usepackage[misc]{ifsym}
\newcommand\blfootnote[1]{\begingroup\renewcommand\thefootnote{}\footnote{#1}\addtocounter{footnote}{-1}\endgroup}

\usepackage[pagebackref=true,breaklinks=true,letterpaper=true,colorlinks,bookmarks=false]{hyperref}

\iccvfinalcopy 

\def\iccvPaperID{6178} 

\begin{document}

\title{Position-Guided Point Cloud Panoptic Segmentation Transformer}

\author{Zeqi Xiao$^{1*}$\quad
Wenwei Zhang$^{1,2*}$\quad
Tai Wang$^{1,3*}$\quad
Chen Change Loy$^{1,2}$\quad
Dahua Lin$^{1,3}$\quad
Jiangmiao Pang$^{1\textrm{\Letter}}$\\
$^1$Shanghai AI Laboratory\quad
$^2$S-Lab, NTU\quad
$^3$The Chinese University of Hong Kong\\
{\tt\small \{zeqixiao1,pangjiangmiao\}@gmail.com, \{wenwei001,ccloy\}@ntu.edu.sg, \{wt019,dhlin\}@ie.cuhk.edu.hk}
}

\maketitle

\blfootnote{* Equal Contribution. \textrm{\Letter} Corresponding author.}








\begin{abstract}
\vspace{-6pt}

DEtection TRansformer (DETR) started a trend that uses a group of learnable queries for unified visual perception.
This work begins by applying this appealing paradigm to LiDAR-based point cloud segmentation and obtains a simple yet effective baseline.
Although the naive adaptation obtains fair results, the instance segmentation performance is noticeably inferior to previous works. 
By diving into the details, we observe that instances in the sparse point clouds are relatively small to the whole scene and often have similar geometry but lack distinctive appearance for segmentation, which are rare in the image domain. 
Considering instances in 3D are more featured by their positional information, we emphasize their roles during the modeling and design a robust Mixed-parameterized Positional Embedding (MPE) to guide the segmentation process. 
It is embedded into backbone features and later guides the mask prediction and query update processes iteratively, leading to Position-Aware Segmentation (PA-Seg) and Masked Focal Attention (MFA).
All these designs impel the queries to attend to specific regions and identify various instances. 
The method, named Position-guided Point cloud Panoptic segmentation transFormer (P3Former), outperforms previous state-of-the-art methods by 3.4\% and 1.2\% PQ on SemanticKITTI and nuScenes benchmark, respectively.
The source code and models are available at \href{https://github.com/SmartBot-PJLab/P3Former}{https://github.com/SmartBot-PJLab/P3Former}.



\end{abstract}

\vspace{-12pt}

\section{Introduction}

\begin{figure}[h!]
  \centering
  \includegraphics[width=\linewidth]{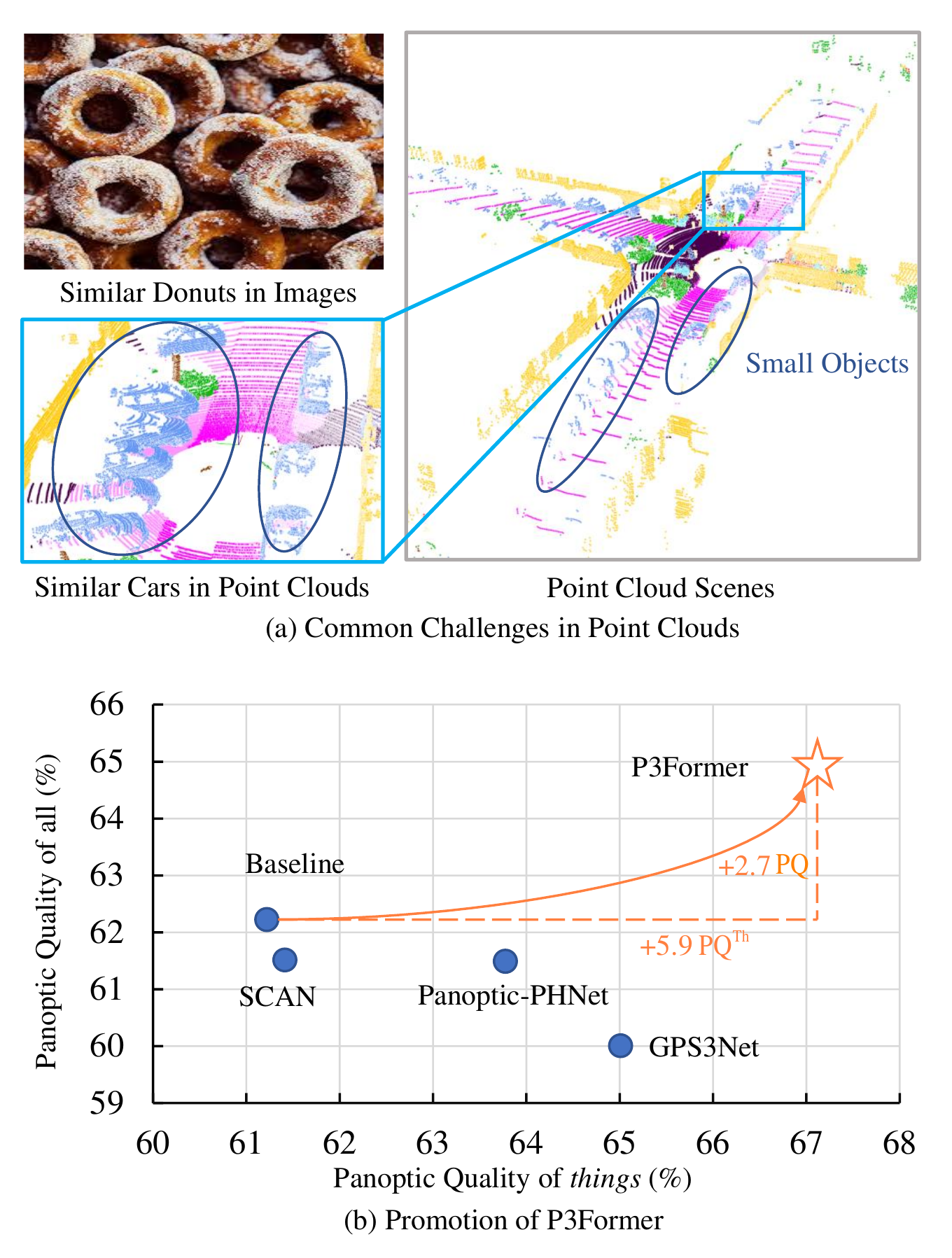}
  \caption{(a) Corner cases in images are common in point cloud scenes: geometrically alike and relatively small objects. (b) Promotion of P3Former with position guidance on SemanticKITTI~\cite{SemanticKITTI} test set, especially for $\mathrm{PQ}^{\mathrm{Th}}$.}
  \label{fig:teaser}
\end{figure}

DEtection TRansformer (DETR)~\cite{DETR} started a trend that uses a group of learnable queries for unified visual perception.
Various methods, such as K-Net~\cite{K-Net} and MaskFormer~\cite{maskformer}, are proposed to unify and simplify the frameworks of image segmentation. 
This work begins by applying this appealing paradigm to LiDAR-based point cloud segmentation and obtains a simple yet effective baseline.
It relies on transformer decoder and bipartite matching for end-to-end training so that each query can attend to different regions of interest and predict the segmentation mask of either a potential object or a stuff class.

This initial attempt unifies and significantly simplifies the framework for LiDAR-based panoptic segmentation.
However, although the naive implementation obtains fair results, the instance segmentation performance is noticeably inferior to previous works~(Fig.~\ref{fig:teaser}-(b)).
Our investigation into this issue reveals two notable challenges in this 3D case: 1) \emph{Geometry ambiguity} (Fig.~\ref{fig:teaser}-(a) left). Instances with similar geometry are much more common in the point clouds than in 2D images and are even harder to be separated due to the lack of texture and colors. 2) \emph{Relatively small objects} (Fig.~\ref{fig:teaser}-(a) right). Instances are typically much smaller with respect to the whole 3D scene, while queries empirically learn to respond to a relatively large area for high recall, making the segmentation of multiple close instances particularly difficult.

Further revisiting the difference between segmentation in images and LiDAR-based point clouds, we observe that instances in images, a dense and structured representation, always have their masks overlapped and their boundaries adhered to each other. In contrast, instances in the 3D space can be clearly separated according to the \emph{positional} information included in their point clouds.
Motivated by this observation, to tackle the aforementioned problems, we fully leverage such properties and propose a Position-guided Point cloud Panoptic Transformer, \emph{P3Former}, which uses a specialized positional embedding to \emph{guide} the whole segmentation procedure.

Specifically, we first devise a \emph{Mixed-parameterized Positional Embedding (MPE)}, which combines Polar and Cartesian spaces. MPE incorporates the Polar pattern prior to the point distribution with the Cartesian embedding, resulting in a robust embedding that serves as the foundation of position-guided segmentation.
It is first embedded into backbone features as the positional discriminator to separate geometrically alike instances.
Furthermore, we also involve it in the mask prediction and masked cross-attention, leading to \emph{Position-Aware Segmentation (PA-Seg)} and \emph{Masked Focal Attention (MFA)}.
\emph{PA-Seg} introduces a parallel branch for position-based mask prediction alongside the original feature-based one. It compensates for the lack of absolute positional information in high-level features.
\emph{MFA} simplifies the masked attention operation by replacing the cross-attention map with our integrated mask prediction.
All these designs enable queries to concentrate on specific positions and predict small masks in a particular region.

We validate the effectiveness of our method on SemanticKITTI~\cite{SemanticKITTI} and nuScenes~\cite{nuScenes} panoptic segmentation datasets.
\emph{P3Former} finally performs even better than previous methods on instance segmentation, resulting in new records with a PQ of 64.9\% on SemanticKITTI and a PQ of 75.9\% on nuScenes, surpassing previous best results by 3.4\% and 1.2\% PQ, respectively (Fig.~\ref{fig:teaser}-(b)).
The simplicity and effectiveness of \emph{P3Former} with our exploration along this pathway shall benefit future research in LiDAR-based panoptic segmentation.


\vspace{-1mm}\section{Related Work}\vspace{-1mm}

\begin{figure*}[t!]
  \centering
  \includegraphics[width=\linewidth]{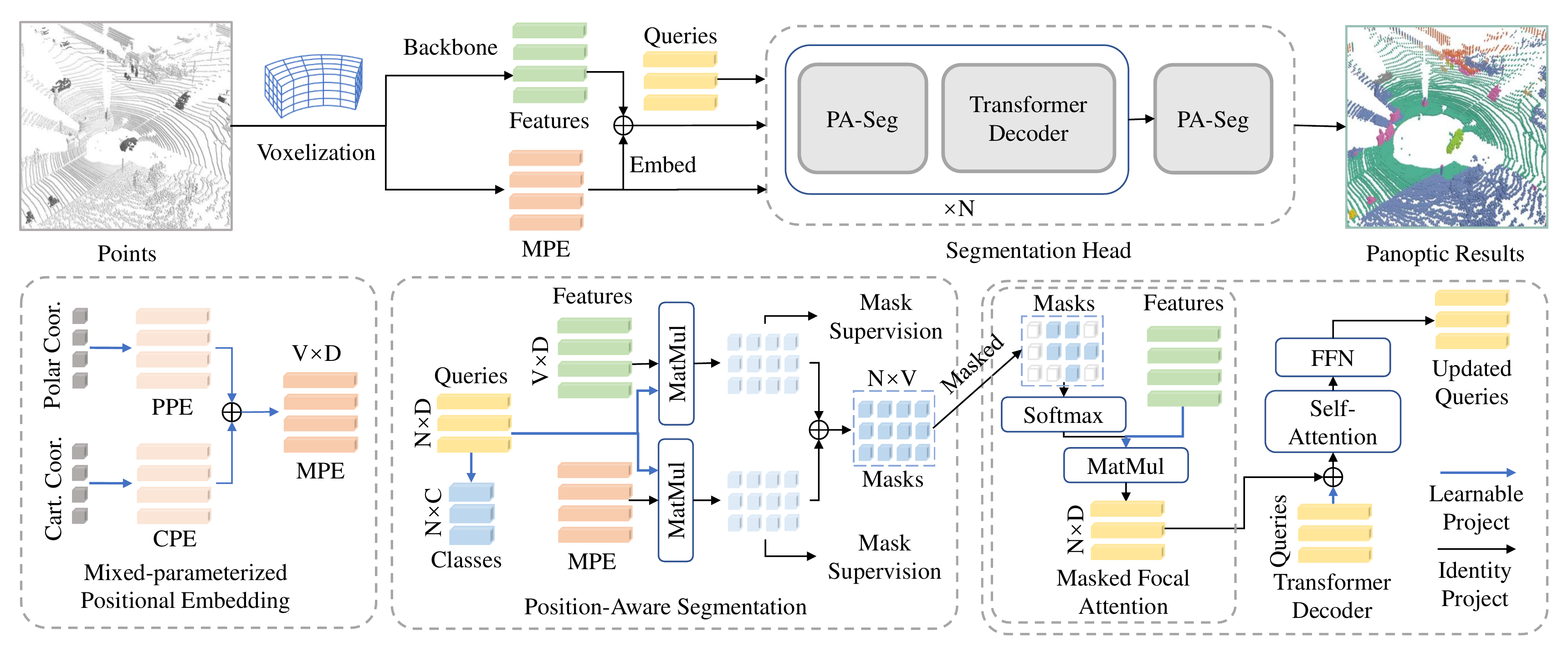}
  \caption{\methodname~ overview. After voxelizing points into voxels, \methodname~uses a backbone to extract voxel features and an MPE module to generate the MPE (Sec. \ref{sec:positional_encoding}). Apart from being embedded into features, MPE further serves as position guidance in the entire segmentation head. In the segmentation head, we use several PA-Seg (Sec. \ref{sec:positional_aware_seg}) and transformer decoder layers to update queries according to features and MPE, then another PA-Seg layer to predict panoptic results with updated queries. The PA-seg consists of two parallel branches to predict masks from features and MPE, with separate supervision. The integrated masks serve as the cross-attention map in Masked Focal Attention (Sec. \ref{sec:mfa}).}
  \label{fig:Overview}
\end{figure*}

\myparagraph{Point Cloud Segmentation.}\quad
Point cloud segmentation aims to map the points into multiple homogeneous groups. Previous works typically resort to different paradigms for indoor~\cite{VV-Net,KPConv,PointConv,LtS,3D-MPA} and outdoor~\cite{Pointnet,Pointnet++, RandLA-Net,Rangenet++,Point-to-Voxel,3D-MiniNet,Cylinder3D,AF2S3Net} scenes, given their difference in the point cloud distribution. Our work focuses on the outdoor case, where the point clouds are obtained by LiDAR and exhibit sparse and non-uniform distribution. The corresponding segmentation problem is called LiDAR-based 3D segmentation.

\noindent\emph{LiDAR-Based 3D semantic segmentation.}\quad
LiDAR-based 3D semantic segmentation categorizes point clouds according to their semantic properties.
Based on different representations of point clouds, we divide existing approaches into three streams, point-based~\cite{Pointnet,Pointnet++,Interpolated}, voxel/grid-based~\cite{RandLA-Net,salsa,Rangenet++,3D-MiniNet,Cylinder3D,AF2S3Net} and multi-modal~\cite{spvnas,rpvnet}. Point-based methods focus on processing individual points, while voxel/grid-based methods quantize point clouds into 3D voxels or 2D grids and apply convolution. Current LiDAR-based segmentation frameworks~\cite{Cylinder3D,AF2S3Net,spvnas} typically adopt voxel-based backbones due to their ability to capture large-scale spatial structures and acceptable computational cost thanks to sparse convolution~\cite{Spconv}. 

\noindent\emph{LiDAR-Based 3D panoptic segmentation.}\quad
Compared with LiDAR-based 3D semantic segmentation, LiDAR-based panoptic segmentation further segments foreground point clouds into different instances.
Most previous top-tier works~\cite{Panoptic-PolarNet,DS-Net,EfficientLPS,GPS3Net,SCAN,PHNet} start from this difference and produce panoptic predictions following a three-stage paradigm, \emph{i.e.}, first predict semantic results, then separate instances based on semantic predictions, and finally fuse the two results.
This paradigm makes the panoptic segmentation performance inevitably bounded by semantic predictions and requires cumbersome post-processing. In contrast, this paper proposes a unified framework with learnable queries for LiDAR segmentation, eliminating all of these problems and predicting panoptic results unanimously.

\myparagraph{Unified Panoptic Segmentation.}\quad
Unified panoptic segmentation was first proposed in the context of 2D image segmentation. Early works built their methods based on either semantic segmentation frameworks~\cite{PanopticPyramid, UPSNet, unifying_CVPR2020, solov2} or instance segmentation frameworks~\cite{panoptic_deeplab, deeplab} to perform panoptic segmentation separately. More recently, DETR~\cite{DETR} simplified the framework for 2D panoptic segmentation, but it still requires two-stage processing. ~\cite{K-Net,maskformer,mask2former} take this a step further by unifying 2D segmentation in one stage. These methods provide excellent solutions for making queries conditional on specific objects, but they require objects to have distinctive features. As a result, these frameworks may be inferior when dealing with objects that have similar texture and geometry. However, such situations are common in LiDAR-based scenes, resulting in ambiguous instance segmentation.

\section{Methodology}\label{sec:methods}

This section elaborates on the detailed formation of P3Former (Sec. Fig.~\ref{fig:Overview}). We start with a panoptic segmentation baseline that simplifies and unifies the cumbersome frameworks in LiDAR-based panoptic segmentation (Sec. \ref{sec:baseline}).
To address the challenges hindering this paradigm from performing well in \emph{things} categories, we propose the Mixed-parameterized Positional Embedding (MPE) to distinguish geometrically alike instances while also fitting the prior distribution of instances (Sec. \ref{sec:positional_encoding}).
Apart from its conventional usage, the MPE provides robust positional guidance for the entire segmentation process. Specifically, it serves as the low-level positional features, paralleling high-level voxel features in position-aware segmentation(Sec. \ref{sec:positional_aware_seg}). Furthermore, 
we replace the masked cross-attention map with the combined masks generated from PA-Seg, making the masked attention focus on small instances while also simplifying the attention structure (Sec. \ref{sec:mfa}).
For more detailed information, please refer to Sec. \ref{sec:implementation}.


\subsection{Baseline}\label{sec:baseline}
The baseline mainly consists of two parts: the backbone and the segmentation head. The backbone converts raw points into sparse voxel features, while the segmentation head predicts panoptic segmentation results through learnable queries~\cite{K-Net,maskformer,mask2former}.

\myparagraph{Backbone.}\label{sec:backbone}
In this work, we choose Cylinder3D~\cite{Cylinder3D} as our backbone for feature extraction due to its good generalization capability for panoptic segmentation~\cite{DS-Net}. 

The backbone takes raw points $P\in \mathbb{R}^{\mathrm{N_P}\times 4}$ ($x,y,z,r$) as input. Through voxelization and sparse convolution, it outputs sparse voxel features $F\in \mathbb{R}^{\mathrm{V}\times \mathrm{D}}$. Here $\mathrm{N_P}$ is the number of points, $r$ is the reflection factor, $\mathrm{V}$ is the number of sparse voxel features, and $\mathrm{D}$ is the feature dimension.

\myparagraph{Segmentation Head.}\label{sec:head}
We follow the backbone to use voxel representation rather than points as the input to the segmentation head to keep the framework efficient. Vanilla unified segmentation heads consist of two components: \emph{One-go Predicting} and \emph{Query updating}. In \emph{One-go Predicting}, each query predicts one object mask and its category, including things and stuff. The process is as follows:
\begin{equation}\label{prediction}
C = f_{C}(Q),\quad M_F = f_{MF}(Q)F^T,
\end{equation}
where $f_{C}$ and $f_{MF}$ are different MLP layer.

To make the predictions more adaptive to specific objects, queries need to be updated according to different inputs.
\emph{Query updating} commonly consists of several transformer-based layers. The inputs of layer $l$ are voxel feature $F$, queries $Q_{l-1}\in \mathbb{R}^{\mathrm{N}\times \mathrm{D}}$, and mask predictions $M_{l-1}\in \mathbb{R}^{\mathrm{V}\times \mathrm{N}}$ from layer $l-1$. The outputs are updated query $Q_{l}$. Here $N$ is the query number. \emph{Query updating} can be formulated as
\begin{equation}
Q_{l} = \mathrm{Updating}(Q_{l-1}, M_{l-1}, F).
\end{equation}
The process can be further decomposed into three parts: Cross-Attention, Self-attention, and Feed-Forward Network (FFN), following the practice of~\cite{mask2former}. These can be generally formulated as
\begin{equation}
Q_{Il} = \mathrm{CrossAttn}(Q_{l-1}, M_{l-1}, F),
\end{equation}
and
\begin{equation}
Q_{l} = \mathrm{FFN}(\mathrm{SelfAttn}(Q_{Il}))),
\end{equation}
respectively. Here we ignore the expression of the $Add\&Norm$ layer for writing simplicity.

The simple and unified framework can obtain competitive results on \emph{stuff} categories compared to previous methods. However, the performance on \emph{things} classes is not ideal due to the special challenges in LiDAR-based scenes.

\subsection{Mixed-parameterized Positional
Embedding}\label{sec:positional_encoding}

\begin{figure}[h!]
  \centering
  \includegraphics[width=\linewidth]{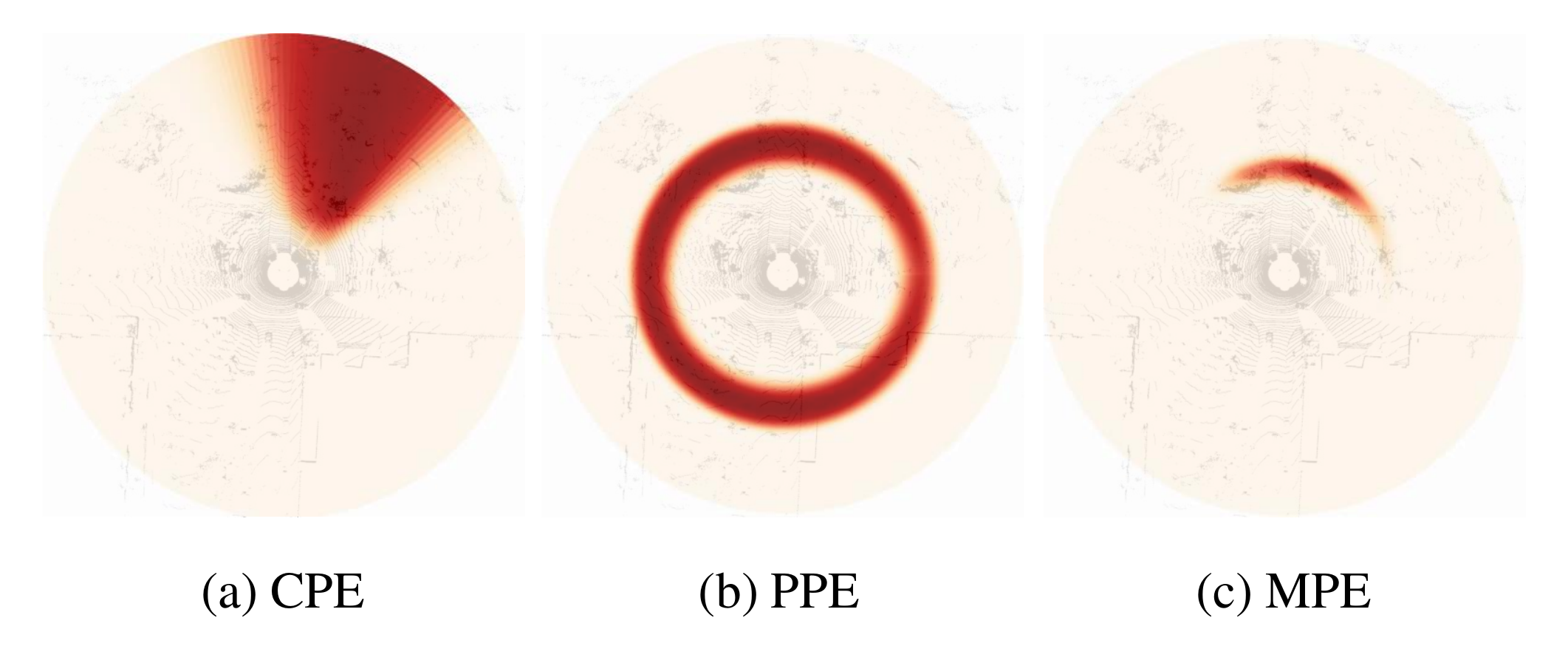}
  \caption{Queries' positional attention distributions of different positional embedding in Bird-Eye-View.}
  \label{fig:pe_cmp}
\end{figure}

Distinguishing instances with the same geometry, texture, and colors is one of the biggest challenges for a unified segmentation structure. As shown in Fig.\ref{fig:teaser}-(a), separating these donuts is one representative hard task for unified segmentation methods in 2D. Unfortunately, such ``donuts" exist nearly everywhere in LiDAR-based scenes: 3D instances don't have information like texture and color. Meanwhile, they are highly geometrically alike if they belong to the same category, e.g., the "car" category shown in Fig.\ref{fig:teaser}-(b). The main indicator we can utilize is their positional difference.

Positional embedding is a good choice to encode positional information into features.
We first formulate positional embedding in Cartesian parameterization, noted as CPE:
\begin{equation}
	\mathrm{CPE} = f_C([x_i, y_i, z_i]).
\end{equation}
Here $f_C$ is a linear transformation layer followed by a normalization layer.
It promisingly enhances instances' discriminability and promotes the performance of the framework. However, this positional embedding parameterization doesn't offer any distribution prior of LiDAR-based scenes. Consequently, to cover instances as many as possible, each query has learned to attend to a big region (Fig.~\ref{fig:pe_cmp}-(a)). While in a crowded scene that contains many relatively small instances, one query would attend to too many instances, which may have negative impacts on the following refinement.

Unlike instances in images that are almost randomly distributed in the picture, instances from LiDAR scans have strong spatial cues (i.e. instances have different distribution patterns depending on distances from the ego vehicle).
Learning the distribution prior is helpful for queries to locate small instances in big scenes. Thus, we design a Polar-parameterized positional embedding to help queries fit the distribution, noted as PPE:
\begin{equation}
	\mathrm{PPE} = f_P([\rho_i, \theta_i, z_i]),
\end{equation}
where $\rho_i = \sqrt{x_i^2+y_i^2}$, $\theta_i = \mathrm{arctan}(y_i/x_i)$, and $f_P$ is also a linear transformation layer followed by a normalization layer. It efficiently helps queries to learn ring-shaped attention (Fig.~\ref{fig:pe_cmp}-(b)).
However, it introduces another problem: instances within the same scan distance are commonly more geometrically alike because of the same point sparsity and scan angle. So they are likely to be segmented as one, although they may not be close in the Cartesian parameterization space.

Based on the observations mentioned above, we propose Mixed Parameterized Positional Embedding (MPE) to leverage the merits of Cartesian and Polar Parameterizations together. Experiments show that simple addition brings ideal promotion above single-format positional embedding. The formulation of MPE is
\begin{equation}
	\mathrm{MPE} = f_P([\rho_i, \theta_i, z_i]) + f_C([x_i, y_i, z_i]),
\end{equation}

We encode MPE into voxel features $F$ and get $F_P$:
\begin{equation}
	F_P = F + MPE,
\end{equation}
which replaces $F$ in the above operations.

\begin{figure}[h]
  \centering
  \includegraphics[width=\linewidth,height=0.75\linewidth]{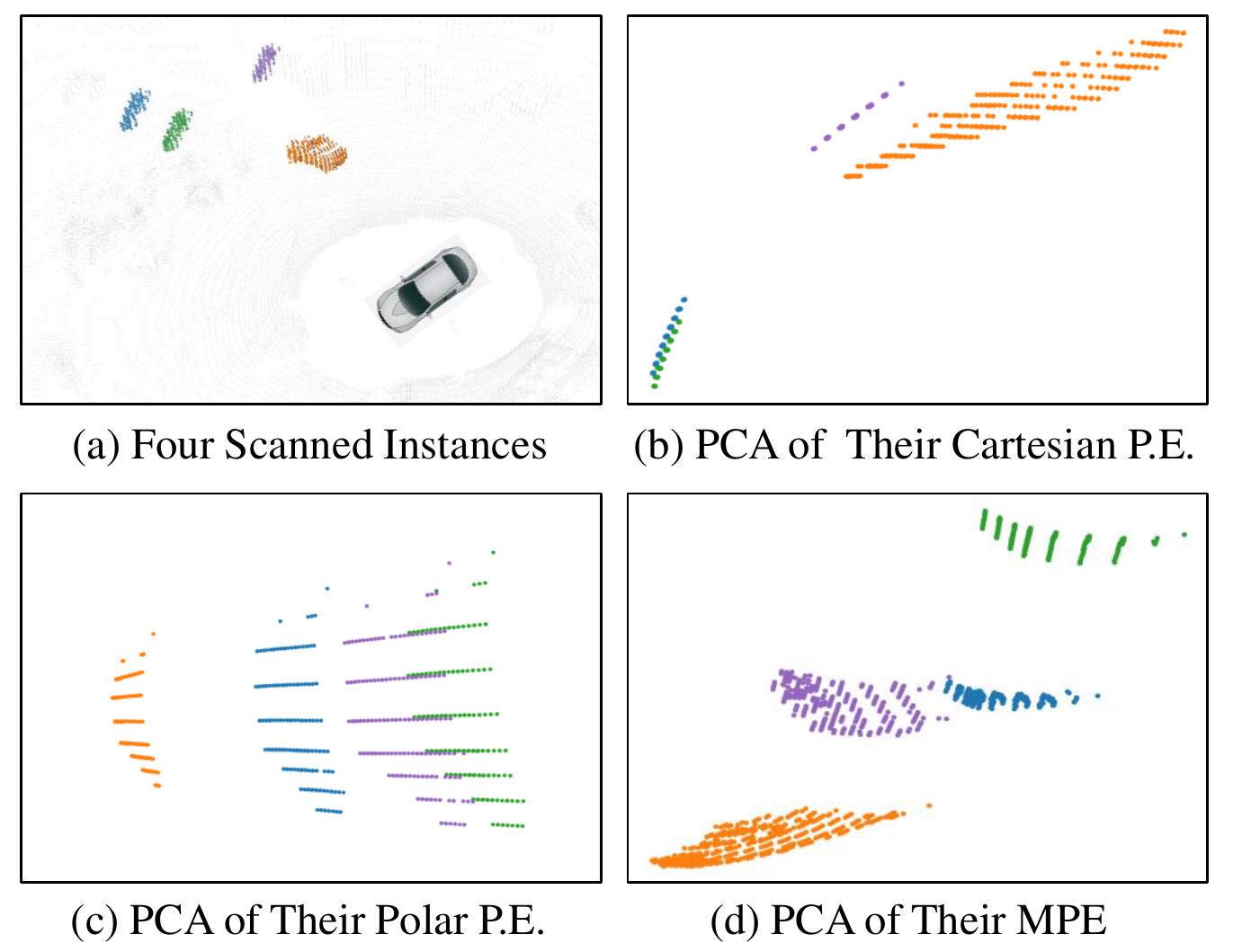}
  \caption{PCA (Principal Component Analysis) of different parameterized positional encoding. Different instances are dyed in origin, green, purple, and blue, respectively. Best viewed in color.}
  \label{fig:mpe}
\end{figure}

Fig.~\ref{fig:pe_cmp}-(c) demonstrates that MPE maintains ring-like distribution prior while also emphasizing the difference in Cartesian space. 
Fig.~\ref{fig:mpe}-(d) shows that MPE successfully pulls away the distance of instances in the MPE space and makes instances more distinguishable. 

Besides, MPE is involved in mask prediction and masked cross-attention, leading to Position-Aware Segmentation (PA-Seg) and Masked Focal Attention (MFA).



\begin{figure}[t]
  \centering
  \includegraphics[width=\linewidth]{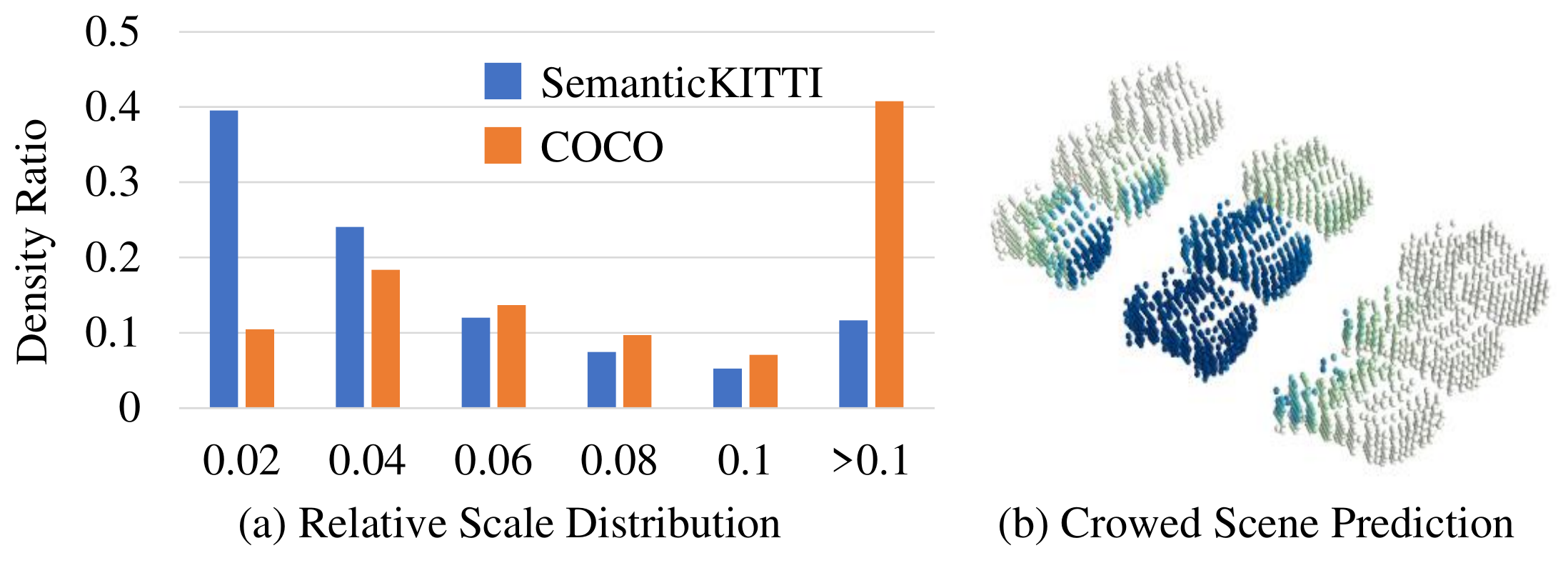}
  \caption{Distribution comparisons of instance relative scale between COCO~\cite{coco} and SemanticKITTI~\cite{SemanticKITTI}. Here instance relative scale= $\sqrt{\mathrm{A_i}/A_s}$ following~\cite{sst}, where $A_i$ is the mask area of one instance, $A_s$ is the mask area of the whole scene.}
  \label{fig:crowd}
\end{figure}

\subsection{Position-Aware Segmentation}\label{sec:positional_aware_seg}

Although the proposed positional embedding strategy can effectively enhance positional information, it is insufficient to deal with small objects, particularly in crowded situations (Fig.~\ref{fig:crowd}-(b)), which are also common in LiDAR-based scenes. We count the relative instances scale distribution in Fig.~\ref{fig:crowd}-(a).
It shows the relative scales of instances in LiDAR-based are much smaller than in images.
This observation requires our segmentation head to be more position-aware to discriminate small instances in a crowd.

We rethink the traditional mask prediction (Eq.~\ref{prediction}) which only depends on voxel features. Vanilla voxel features generated from the hourglass-like backbone mainly capture high-level geometric information and lack low-level positional information. The embedded MPE partially compensates for the drawback.
However, such an operation just uses the MPE implicitly. To this end, we directly predict masks from MPE so queries can explicitly acquire low-level information about a specific position. The formulation is
\begin{equation}\label{MPE_prediction}
	M_P = f_{MP}(Q)\cdot \mathrm{MPE},
\end{equation}
where $M_P$ is position mask and $f_{MP}$ is a MLP layer.

It's worth noting that we supervise $M_F$ and $M_P$ separately. We simply use the binary ground truth instance masks $G_M$ as the learning targets of $M_P$. This is feasible because, in LiDAR-based scenes, instances are concentrated in small local regions and don't overlap with each other. It shares the same function of center heatmaps used in previous methods, while our solution is more succinct and doesn't need to generate pseudo centers of instances.

We add $M_F$ and $M_P$ together to get the final masks $M$, which contain both high-level geometric and low-level positional information. Experiments show that queries with explicit positional awareness will quickly converge to a local region and focus on a small instance.

\begin{table*}[t]
  \footnotesize
  \begin{center}
  \caption{\centering Comparisons of LiDAR panoptic segmentation performance on SemanticKITTI test split. * We don't use TTA for a fair comparison.}
  \label{tab:benchmark:SemanticKITTI:test}
    \scalebox{0.90}{\tablestyle{8pt}{1.0}
    \begin{tabular}{l|p{0.4cm}<{\centering}p{0.4cm}<{\centering}p{0.4cm}<{\centering}p{0.6cm}<{\centering}|p{0.4cm}<{\centering}p{0.4cm}<{\centering}p{0.6cm}<{\centering}|p{0.4cm}<{\centering}p{0.4cm}<{\centering}p{0.6cm}<{\centering}p{0.6cm}<{\centering}}
    
    Methods & PQ & $\mathrm{PQ^{\dag}}$ &RQ & SQ & $\mathrm{PQ^{Th}}$ & $\mathrm{RQ^{Th}}$ & $\mathrm{SQ^{Th}}$ & $\mathrm{PQ^{St}}$ & $\mathrm{RQ^{St}}$ & $\mathrm{SQ^{St}}$ & FPS\\
    
    \hline
    \specialrule{0.05em}{3pt}{3pt}
    RangeNet++~\cite{Rangenet++}/PointPillars~\cite{PointPillars} & 37.1 & 45.9 & 75.9 & 47.0 & 20.2 & 75.2 & 25.2 & 49.3 & 62.8 & 76.5 & 2.4\\
    LPSAD~\cite{LPSAD}                   & 38.0 & 47.0 & 48.2 & 76.5 & 25.6 & 31.8 & 76.8 & 47.1 & 60.1 & 76.2 & 11.8\\
    KPConv~\cite{KPConv}/PointPillars~\cite{PointPillars} & 44.5 & 52.5 & 54.4 & 80.0 & 32.7 & 38.7 & 81.5 & 53.1 & 65.9 & 79.0 & 1.9\\    
    Panoster~\cite{Panoster}                & 52.7 & 59.9 & 64.1 & 80.7 & 49.4 & 58.5 & 83.3 & 55.1 & 68.2 & 78.8 & -\\
    Panoptic-PolarNet~\cite{Panoptic-PolarNet}       & 54.1 & 60.7 & 65.0 & 81.4 & 53.3 & 60.6 & 87.2 & 54.8 & 68.1 & 77.2 & 11.6\\
    DS-Net~\cite{DS-Net}                  & 55.9 & 62.5 & 66.7 & 82.3 & 55.1 & 62.8 & 87.2 & 56.5 & 69.5 & 78.7 & 2.1\\
    EfficientLPS~\cite{EfficientLPS}            & 57.4 & 63.2 & 68.7 & 83.0 & 53.1 & 60.5 & 87.8 & 60.5 & 74.6 & 79.5 & - \\
    GP-S3Net~\cite{GPS3Net}                & 60.0 & 69.0 & 72.1 & 82.0 & 65.0 & 74.5 & 86.6 & 56.4 & 70.4 & 78.7 & - \\
    SCAN~\cite{SCAN}                    & 61.5 & 67.5 & 72.1 & 84.5 & 61.4 & 69.3 & 88.1 & 61.5 & 74.1 & 81.8 & 12.8\\
    Panoptic-PHNet~\cite{PHNet}         & 61.5 & 67.9 & 72.1 & 84.8 & 63.8 & 70.4 & 90.7 & 59.9 & 73.3 & 80.5 & 11.0  \\
    \hline
    \specialrule{0.05em}{3pt}{3pt}
    \methodname             & \textbf{64.9} & \textbf{70.0} & \textbf{75.9} & \textbf{84.9} & \textbf{67.1} & \textbf{74.1} & \textbf{90.6} & \textbf{63.3} & \textbf{77.2} & \textbf{80.7} & 11.6\\
    \end{tabular}}
  \end{center}
\end{table*}  

\begin{table*}[t]
  \footnotesize
  \begin{center}
  \caption{Comparisons of LiDAR panoptic segmentation performance on nuScenes dataset}
  \label{tab:benchmark:nuScenes} 
  \scalebox{0.90}{\tablestyle{8pt}{1.0}
    \begin{tabular}{l|p{0.4cm}<{\centering}p{0.4cm}<{\centering}p{0.4cm}<{\centering}p{0.6cm}<{\centering}|p{0.4cm}<{\centering}p{0.4cm}<{\centering}p{0.6cm}<{\centering}|p{0.4cm}<{\centering}p{0.4cm}<{\centering}p{0.6cm}<{\centering}}
    
    Methods & PQ & $\mathrm{PQ^{\dag}}$ &RQ & SQ & $\mathrm{PQ^{Th}}$ & $\mathrm{RQ^{Th}}$ & $\mathrm{SQ^{Th}}$ & $\mathrm{PQ^{St}}$ & $\mathrm{RQ^{St}}$ & $\mathrm{SQ^{St}}$\\
    \hline
    \specialrule{0.05em}{3pt}{3pt}    

    DS-Net~\cite{DS-Net}         & 42.5 & 51.0 & 83.6 & 50.3 & 32.5 & 38.3 & 83.1 & 59.2 & 84.4 & 70.3 \\
    EfficientLPS~\cite{EfficientLPS}   & 59.2 & 62.8 & 82.9 & 70.7 & 51.8 & 62.7 & 80.6 & 71.5 & 84.1 & 84.3 \\
    GP-S3Net~\cite{GPS3Net}       & 61.0 & 67.5 & 72.0 & 84.1 & 56.0 & 65.2 & 85.3 & 66.0 & 78.7 & 82.9 \\
    SCAN~\cite{SCAN}           & 65.1 & 68.9 & 75.3 & 85.7 & 60.6 & 70.2 & 85.7 & 72.5 & 83.8 & 85.7 \\
    Panoptic-PHNet~\cite{PHNet} & 74.7 & 77.7 & 84.2 & 88.2 & 74.0 & 82.5 & 89.0 & 75.9 & 86.9 & 76.5  \\
    \hline
    \specialrule{0.05em}{3pt}{3pt}
    \methodname                 & \textbf{75.9} & \textbf{78.9} & \textbf{84.7} & \textbf{89.7} & \textbf{76.9} & \textbf{83.3} & \textbf{92.0} & 75.4 & \textbf{87.1} & \textbf{86.0}\\
    \end{tabular}}
  \end{center}
\end{table*}  

\subsection{Masked Focal Attention}\label{sec:mfa}
The integrated mask $M$, which is augmented with position-focusing awareness, can be leveraged as the cross-attention map for query-feature interaction.
Vanilla segmentation heads adopt \emph{Masked Cross Attention}, which is
\begin{equation}\label{MaskCrossAttention}
	Q_{Il} = \mathrm{softmax}(A_{l-1} + q_lk_l^T)v_l + Q_{l-1},
\end{equation}
where $q_l = f_q(Q_{l-1})$, $k_l = f_k(F)$, $v_l = f_v(F)$. $f_q(\cdot), f_k(\cdot), f_v(\cdot)$ are all linear transformations. The attention mask $A_{l-1}$ at feature location $v \in V$ is
\begin{equation}
A_{l-1}(v)=\left\{
\begin{array}{rcl}
	    0 & & if\ M_{l-1}(v) > 0,\\
	    -\infty & & otherwise.
\end{array}
\right.
\end{equation}

Here the attention map generated by $q_lk_l^T=f_q(Q_l)f_v(F)^T$ resembles mask prediction $M_F$ in Eq.~\ref{prediction} which only uses high-level features $F$. Inspired by Sec.~\ref{sec:positional_aware_seg}, we can leverage masks $M$ from position-aware segmentation as the attention map.
So Eq.~\ref{MaskCrossAttention} can be reformulated to
\begin{equation}
	Q_{Il} = \mathrm{softmax}(A_{l-1} + M)v_l + Q_{l-1}.
\end{equation}
We refer to it as \emph{Masked Focal Attention} since the attention features inherit the merit of position-aware masks and focus more on small instances.

\subsection{Training and Inference}\label{sec:implementation}
During training, since the number of predictions $N$ is larger than the pre-defined number of classes, we adopt bipartite matching and set prediction loss to assign ground truth to predictions with the smallest matching cost. For inference, we simply use \emph{argmax} to determine the final panoptic results.

\myparagraph{Loss Functions. }
Our loss function is composed of a classification loss, a feature-seg loss, and a position-seg loss. We choose focal loss~\cite{focalloss}~$l_{f}$ for the classification loss $L_{c}$. The feature-seg loss $L_{fs}$ is a weighted summation of binary focal loss $l_{bf}$ and dice loss~\cite{dice_loss}~$l_{df}$. We use dice loss $l_{fp}$ for the position-seg loss $L_{ps}$. The full loss function $L$ can be formulated as $L=L_{c}+L_{fs}+L_{ps}$, while $L_{c}=\lambda_{f}l_{f}$, $L_{fs}=\lambda_{bf}l_{bf}+\lambda_{df}l_{df}$, $L_{ps}=\lambda_{fp}l_{fp}$. In our experiments, we empirically set $\lambda_{f}:\lambda_{bf}:\lambda_{df}:\lambda_{fp}=1:1:2:0.2$.

\myparagraph{Hungarian Assignment. }
Following~\cite{DETR,maskformer,K-Net}, we adopt the Hungarian assignment strategy to build one-to-one matching between our predictions and ground truth. The formation of matching cost is the same as our loss function.

\myparagraph{Inference. }
Unlike previous LiDAR-based panoptic segmentation methods that need manually designed fusion strategies on semantic and instance segmentation results, we only need to perform \emph{argmax} among $N$ mask predictions to get the final panoptic results with the dimensions of $1\times \mathrm{V}$.

\myparagraph{Iterative Refinement. }
We stack the Transformer layer several times to refine learnable queries in our framework. During training, we supervise the results predicted by queries before interaction and after each refinement layer with our loss functions. The classification loss is not required for queries before interaction because, at that time, learnable queries are not conditional on specific semantic classes. For inference, we use the results predicted by the finally refined queries.

\section{Experiments}\label{sec:Experiments}
In this section, we present the experimental setting and benchmark results on two popular LiDAR-based panoptic segmentation datasets: SemanticKITTI~\cite{SemanticKITTI} and nuScenes~\cite{nuScenes}. We also ablate the detailed designs of our proposed methods and provide representative visualizations for qualitative analysis.

\subsection{Experimental Setting}\label{sec:setting}

\myparagraph{SemanticKITTI. }
SemanticKITTI~\cite{SemanticKITTI,semantickittipan} is the first large-scale dataset on LiDAR-based panoptic segmentation. 
It contains 43552 frames of outdoor scenes, of which 23201 frames with panoptic labels are used for training and validation, and the remaining 20351 frames without labels are used for testing. 
The annotations include 8 things classes and 11 stuff classes out of 19 semantic classes.

\myparagraph{nuScenes.}
nuScenes~\cite{fong2021panoptic} is a public autonomous driving dataset. It provides a total of 1000 scenes, including 850 scenes (34149 frames) for training and validation and 150 scenes (6008 frames) for testing. Among the 16 semantic classes, there are 10 things classes and 6 stuff classes.

\myparagraph{Evaluation Metrics.}
We use the panoptic quality (PQ)~\cite{panoptic} as our main metric to evaluate the performance of panoptic segmentation. PQ can be seen as the multiplication of segmentation quality (SQ) and recognition quality (RQ), which is formulated as

\eqnsm{psq-seg-det}{\small{\text{PQ}} = \underbrace{\frac{\sum_{\TP} \text{IoU}}
{|\TP|}}_{\text{SQ}} \times \underbrace{\frac{|\TP|}{|\TP| + \frac{1}{2} |\FP| + \frac{1}{2} |\FN|}}_{\text{RQ}}.}

These three metrics can be extended to things and stuff classes, denoted as $\mathrm{PQ^{Th}}$, $\mathrm{PQ^{St}}$, $\mathrm{RQ^{Th}}$, $\mathrm{RQ^{St}}$, $\mathrm{SQ^{Th}}$, $\mathrm{SQ^{St}}$, respectively. We also report PQ$^{\dagger}$ proposed by~\cite{PQdagger}, which replaces PQ with $\text{IoU}$ for stuff classes.

\myparagraph{Implementation Details.}
We implement \methodname~with MMDetection3D~\cite{mmdet3d}. We follow Cylinder3D's~\cite{Cylinder3D} procedures for sparse feature extraction and data augmentation on both SemanticKITTI and nuScenes. For sparse feature extraction, we discretize the 3D space to $ 480\times360\times32$ voxels.
For data augmentation, we adopt rotation, flip, scale, and noise augmentation. We choose AdamW~\cite{ADAMW} as the optimizer with a default weight decay of 0.01. In the ablation study, we train the model for 40 epochs with a batch size of 4 on 4 NVIDIA A100 GPUs. The initial learning rate is 0.005 and will decay to 0.001 in epoch 26. More details are provided in the appendix.

\subsection{Benchmark Results}

\myparagraph{SemanticKITTI.}
We compare our method with RangeNet++~\cite{Rangenet++} + PointPillars~\cite{PointPillars}, LPSAD~\cite{LPSAD}, KPConv~\cite{KPConv} + PointPillars~\cite{PointPillars}, Panoster~\cite{Panoster}, Panoptic-PolarNet~\cite{Panoptic-PolarNet}, DS-Net~\cite{DS-Net}, EfficientLPS~\cite{EfficientLPS}, GP-S3Net~\cite{GPS3Net}, SCAN~\cite{SCAN} and Panoptic-PHNet\cite{PHNet}. Table~\ref{tab:benchmark:SemanticKITTI:test} shows comparisons of LiDAR panoptic segmentation performance on the SemanticKITTI test split. Our method surpasses the best baseline method by 3.4\% in terms of PQ. Notably, we outperform the second method ~\cite{PHNet} by 3.3\% and 3.4\% in $\mathrm{PQ^{Th}}$ and $\mathrm{PQ^{St}}$, respectively, suggesting that our unified framework gains in both things and stuff classes.

\myparagraph{nuScenes.}
We also provide the comparison results of LiDAR panoptic segmentation performance on nuScenes validation split. Compared with SemanticKITTI, the point clouds contained in scenes of nuScenes are more sparse, so it can better highlight the model's ability to model long-distance information. We compare our method with DS-Net~\cite{DS-Net}, EfficientLPS~\cite{EfficientLPS}, GP-S3Net~\cite{GPS3Net}, SCAN~\cite{SCAN} and Panoptic-PHNet~\cite{PHNet}. As shown in Table~\ref{tab:benchmark:nuScenes}, our method outperforms existing methods on most of the metrics, surpassing the runner-up method by 1.2\% and 2.9\% in terms of PQ and $\mathrm{PQ^{Th}}$. It shows that our method has a strong modeling ability for long-distance information and maintains a strong segmentation ability for sparse instances.

\begin{figure}[t]
  \centering
  \includegraphics[width=\linewidth]{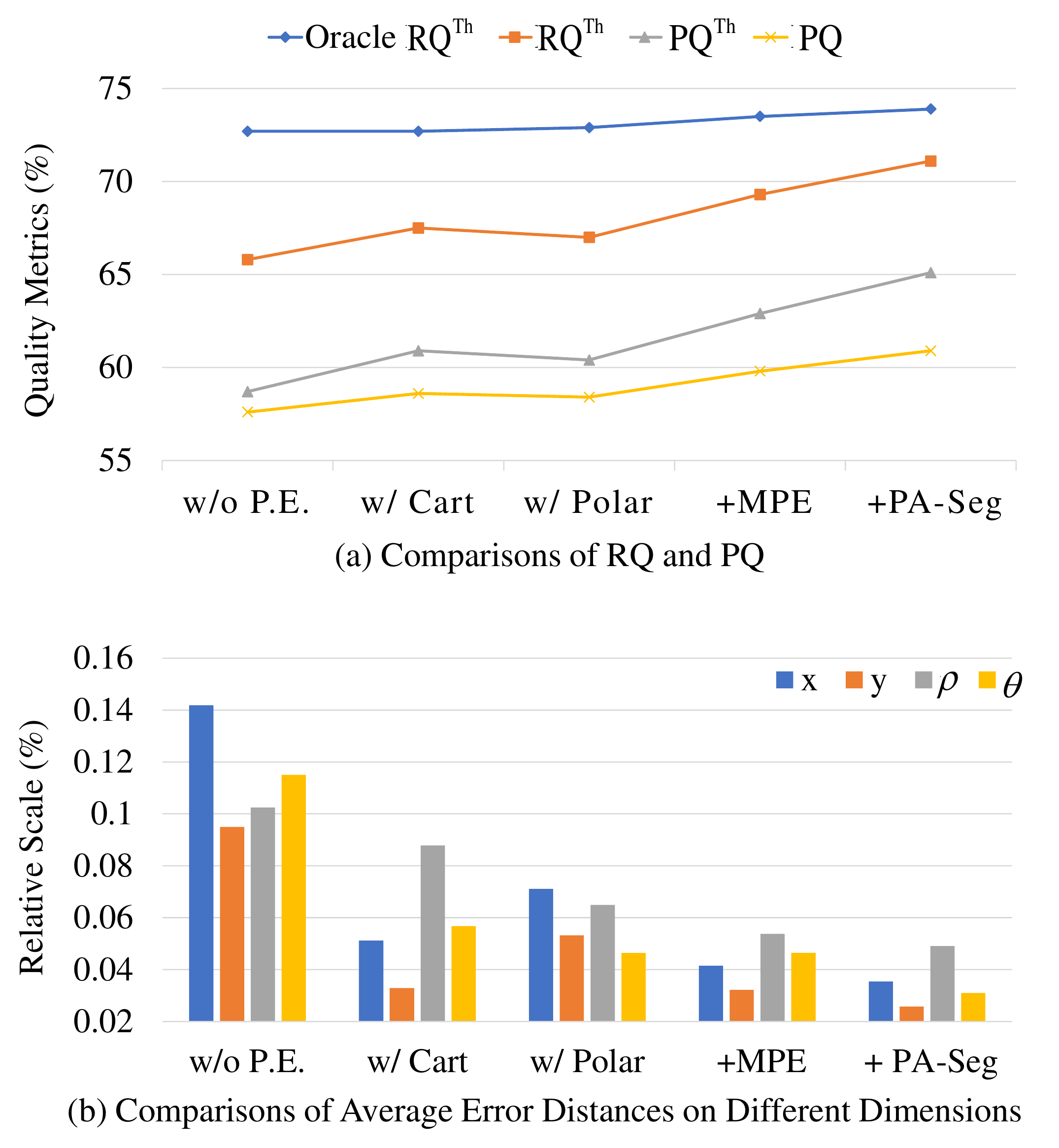}
  \caption{Comparisons of different designs in our model. ``Oracle $\mathrm{RQ^{Th}}$'' takes all instances included in things prediction as $\TP$. ``Error distances" measures the one-to-one distances of instances being wrongly predicted as one.}
  \label{fig:comparison_of_many}
\end{figure}

\begin{figure*}[t]
  \centering
  \includegraphics[width=\linewidth]{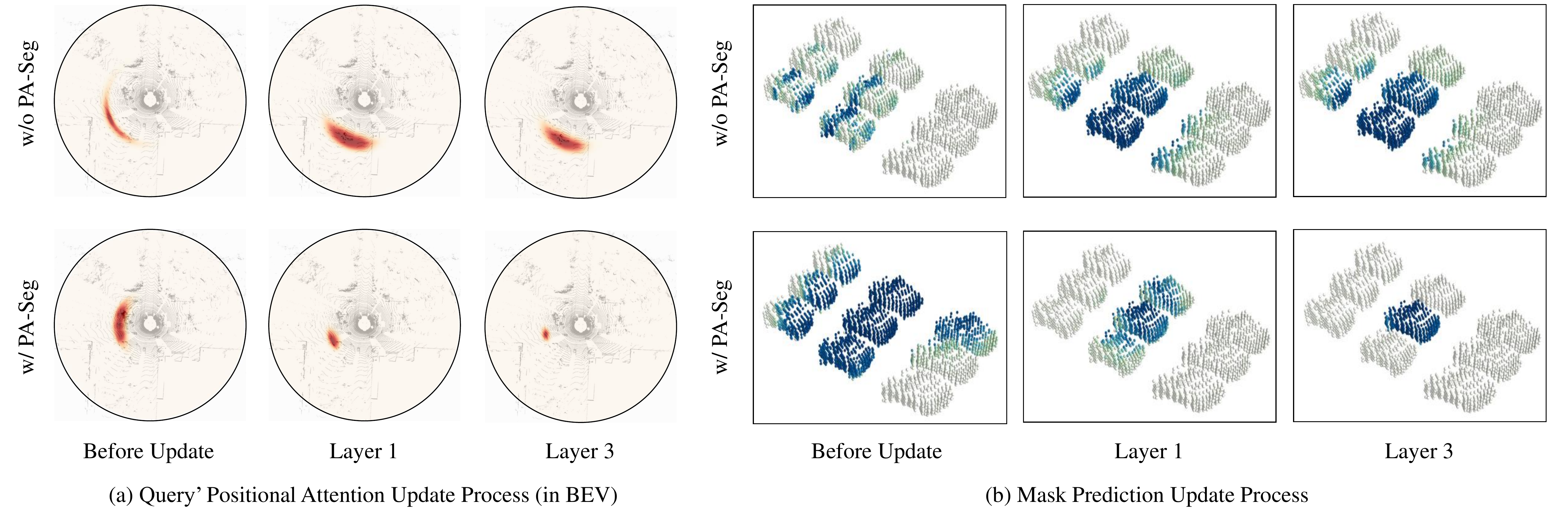}
  \caption{Comparisons of update process with and without PA-Seg. (a) Darker red indicates a high attention level.  (b) We construct a crowed scene including 9 similar cars and visualize one query's mask prediction. Dark blue indicates high prediction confidence.}
  \vspace{-12pt}
  \label{fig:attn_update}
\end{figure*}

\subsection{Ablation Studies}
We conduct several groups of ablations studies in this section to demonstrate the effectiveness of~\methodname~and detailed information about the framework. All experiments are based on SemanticKITTI~\cite{SemanticKITTI} validation split.

\myparagraph{Effects of Components.}
We ablate each component that improves the performance of~\methodname~in Table~\ref{tab:ablation:components}. Our proposed MPE and PA-Seg promote 1.8\% PQ (2.8\% $\mathrm{PQ^{Th}}$) and 1.1\% PQ (2.7\% $\mathrm{PQ^{Th}}$) above baseline, respectively. MFA further improve 0.5\% PQ and 1.4\% $\mathrm{PQ^{Th}}$.

\begin{table}[h]
  \begin{center}
  \caption{Ablation studies of~\methodname~on SemanticKITTI validation set}
  \label{tab:ablation:components}
  \scalebox{0.90}{\tablestyle{8pt}{1.0}
    \begin{tabular}{cccc|p{0.6cm}<{\centering}p{0.6cm}<{\centering}p{0.9cm}<{\centering}p{0.6cm}<{\centering}p{0.6cm}<{\centering}p{0.6cm}}
    
    Baseline & MPE & PS & MFA & PQ & $\mathrm{PQ^{Th}}$ & $\mathrm{PQ^{St}}$\\
    \hline
    \specialrule{0.05em}{3pt}{3pt}
      \checkmark  &                &            &            &            58.0 & 59.6 & 57.0\\
      \checkmark  &   \checkmark   &            &            &            59.8 & 62.4 & 57.9\\
      \checkmark  &   \checkmark   & \checkmark &            &            60.9 & 65.1 & 57.9\\
      \checkmark  &   \checkmark   & \checkmark & \checkmark &            61.4 & 66.5 & 57.7\\
    \end{tabular}}
  \end{center}
  \vspace{-12pt}
\end{table}

\myparagraph{Ambiguous Instance Segmentation.}
\emph{Ambiguous Instance Segmentation} (AIS) refers to the case that one query segments multiple instances due to the close feature representation.
We conduct a series of experiments to prove that AIS is one of the salient bottlenecks in this task.
Specifically, we start from the metric $\mathrm{RQ^{Th}}$ since it is a key component of $\mathrm{PQ^{Th}}$ and a direct indicator of the model's capability to locate instances. 
As shown in Fig.~\ref{fig:comparison_of_many}-(a), there is a huge gap between Oracle $\mathrm{RQ^{Th}}$ and $\mathrm{RQ^{Th}}$ for baseline setting. Since Oracle $\mathrm{RQ^{Th}}$ ignores the AIS cases, it indicates that there are considerable numbers of AIS cases due to the lack of appearance information. Our proposed ingredients effectively remedy the gap, accompanied by the promotion of performance. Here the formulation of Oracle $\mathrm{RQ^{Th}}$ is the same as $\mathrm{RQ^{Th}}$, while we reformulate $\text{IoU}$ as $\text{Oracle IoU}$:
\begin{equation}
    \text{Oracle IoU} = \frac{P\cap T}{T},
\end{equation}
where $P$ indicates prediction areas and $T$ indicates target areas. $\text{Oracle IoU}$ measures the percentage of areas that are covered by predictions.

\vspace{12pt}

\myparagraph{Mixed-parameterized Positional Encoding (MPE). }\label{sec:MPE}
We compare different types of positional encoding in Fig.~\ref{fig:comparison_of_many}. Fig.~\ref{fig:comparison_of_many}-(a) shows that Cartesian parameterization (Cart) and Polar parameterization (Polar) are both effective in terms of PQ. The combination of these two can further bring improvements. In Fig.~\ref{fig:comparison_of_many}-(b) we gather the statistics of the average distances of instances that are wrongly segmented as one on four dimensions $(x,y,\theta,\rho)$. Since instances with smaller distances are harder to separate, methods that achieve smaller average distances on a specific dimension are more accurate on that dimension. It proves that single-parameterized positional encoding can partially improve the model's accuracy on some dimensions but is deficient on others. A mixed-parameterized one can remedy this drawback.

\begin{table}[h]
    \caption{Ablation of Segmentation by Position (SP). ``RS" stands for ``Relative Scale", measuring the relative scale of mask predictions.}
    \vspace{-12pt}
    \begin{center}
    \small{
    \begin{tabular}{c|c|ccc}
    Metrics             & PA-Seg        & Layer 0 & Layer 1 & Layer 3\\
    \hline
    \specialrule{0.05em}{3pt}{3pt}
    RS (\%)             &            & 0.146 & 0.062 & 0.036\\
                        & \checkmark & 0.136 & 0.045 & 0.030\\
    \specialrule{0.05em}{3pt}{3pt}
    PQ (\%)             &            & 55.8 & 58.3 & 59.8 \\
                        & \checkmark & 56.0 & 59.8 & 60.9 \\
    \specialrule{0.05em}{3pt}{3pt}
    $\mathrm{PQ^{Th}}$ (\%) &        & 54.5 & 60.1 & 62.9\\
                        & \checkmark & 54.6 & 63.0 & 65.1 \\
    \end{tabular}
        }
    \end{center}
    \label{tab:pf_loss}
    \vspace{-12pt}
\end{table}

\myparagraph{Position-Aware Segmentation (PA-Seg).}
We validate the effectiveness of PA-Seg in many aspects. Fig.~\ref{fig:comparison_of_many}-(a) shows that PA-Seg reduces the AIS cases and closes the gap of Oracle $\mathrm{RQ^{Th}}$ and $\mathrm{RQ^{Th}}$. With queries specialize in focused positions, their mask predictions shrink to a more precise area, resulting in smaller relative scale (Table~\ref{tab:pf_loss}) of mask prediction and smaller average error distances (Fig.~\ref{fig:comparison_of_many}-(b)). PA-Seg leads the queries to quickly be adaptive to their corresponding objects, with larger $\mathrm{PQ}$ and $\mathrm{PQ^{Th}}$ promotion during iterative refinements (Table~\ref{tab:pf_loss}).



\subsection{Visual Analysis}\label{sec:Analysis}

Fig~\ref{fig:attn_update}-(a) illustrates the update process of query positional attention with and without PA-Seg. Queries initially learn positional priors from the dataset and pay attention to a ring-shaped region. During the update, queries without PA-Seg are not aware of updating their positional attention. However, with PA-Seg, queries quickly converge their positional attention and specialize in a small region. Fig.~\ref{fig:attn_update}-(b) further demonstrates the impact of PA-Seg on mask predictions. The specialization of positional focal attention enabled by PA-Seg aids in successfully separating a car from a crowded scene.


\vspace{-1mm}\section{Conclusion}\vspace{-1mm}

We present \methodname~for unified LiDAR-based panoptic segmentation. 
It builds on the DETR paradigm and addresses the challenges of distinguishing small and geometrically similar instances. 
We introduce a robust Mixed-parameterized Positional Embedding which is embedded into voxel features and further guides the segmentation process, leading to Position-Aware Segmentation and Masked Focal Attention. All these designs enable queries to concentrate on specific positions and predict small masks in a particular region.
State-of-the-art performance on SemanticKITTI and nuScenes demonstrate the effectiveness of our framework and reveal the potential of \methodname~with more robust 3D representations in future research.


\clearpage

\appendix
\begin{center}{\bf \Large Supplementary Material}\end{center}\vspace{-1mm}

\vspace{-1mm}\section{Implementation Details}\label{sec:appendix:implementation}\vspace{-1mm}

For data augmentation, we adopt rotating, flipping, scaling, and noising.

In the voxelization process, we first transform raw point clouds' Cartesian coordinates ($x,y,z$) into Polar coordinates ($r, \theta, z$). Then we clip the 3D space range into $r\in [0, 50], \theta\in [-\pi, \pi], z\in [-4, 2]$ for SemanticKITTI~\cite{SemanticKITTI}, and $r\in [0, 50], \theta\in [-\pi, \pi], z\in [-5, 3]$ for nuScenes~\cite{nuScenes}.

For SemanticKITTI~\cite{SemanticKITTI} test split results, we use both training split and validation split for training. We train the model for 80 epochs with a batch size of 8 on eight NVIDIA A100 GPUs, as more epochs can continuously optimize the performance. The initial learning rate is 0.005 and will decay to 0.001 in epoch 60.

For inference, we keep mask predictions with sigmoid classification confidences larger than 0.4. After $argmax$, we filter out mask predictions that the IoU of the remaining mask and original mask is smaller than 0.8.

\vspace{-1mm}\section{Discussions}\label{sec:appendix:discussion}\vspace{-1mm}

\myparagraph{Model Comparisons.}
We compare panoptic segmentation models based on the Cylinder3D~\cite{Cylinder3D} backbone. As shown in Table~\ref{tab:setting}. We outperform previous methods~\cite{Cylinder3D, DS-Net} by a large margin in performance and speed. The comparisons of~\methodname~with different settings demonstrate that while heavier network settings promote performance, they also add on computational cost and lower the inference speed.

\myparagraph{Point vs. Voxel.}
To implement P3Former, an important choice is the type of 3D representations for segmentation. P3Former uses sparse voxel features rather than points as the input of the segmentation head. Table~\ref{tab:PV} proves that the voxel type outperforms the point type by 1.5 PQ while about 1.7 times faster in inference. The performance gain of the voxel type is intuitive because, by voxelizing points, we rebalance the varying point density of objects in the scenes.

\begin{table}[h]
    \caption{Ablation of Representation Types.}
    \begin{center}
    \small{
    \begin{tabular}{c|c|c}
      Feature Type & PQ & FPS\\
    \hline
    \specialrule{0.05em}{3pt}{3pt}
      Voxel  &  60.9 & 19.7\\
      Point  &  59.4 & 11.4\\
     \end{tabular}
        }
    \end{center}
    \label{tab:PV}
\end{table}

\myparagraph{Semantic Segmentation.}
We verify the effectiveness of \methodname~on semantic segmentation of SemanticKITTI (Table~\ref{tab:sem_seg}). With interaction mechanism and iterative refinement, we outperform vanilla Cylinder3D decode head by 1.7\% mIOU. 

\begin{table}[h]
    \caption{Results of Semantic Segmentation}
    \begin{center}
    \small{
    \begin{tabular}{c|c}
    Method & mIOU\\
    \hline
    \specialrule{0.05em}{3pt}{3pt}
    Cylinder3D & 62.5 \\
    \methodname & 64.2 \\
    \end{tabular}
        }
    \end{center}
    \label{tab:sem_seg}
\end{table}

\myparagraph{Loss Weights.}
Objects are relatively tiny compared with the whole outdoor scene. Thus, the proportion of positive and negative samples in mask supervision is critically imbalanced. From this observation, we select Focal Loss~\cite{focalloss} rather than CrossEntropy Loss to alleviate this problem. Besides, we increase the weight of Dice Loss since it can supervise the mask without being influenced by the imbalanced situation. The experimental results are shown in Table~\ref{tab:ablation:loss}.

\begin{table}[h]
    \caption{Ablation of Loss Weights.}
    \begin{center}
    \small{
    \begin{tabular}{ccc|ccc}
     CE & Focal & Dice & PQ & RQ & SQ\\
    \hline
    \specialrule{0.05em}{3pt}{3pt}
    1 & 0 & 1 & 58.9 & 68.9 & 75.0\\
    0 & 1 & 1 & 59.0 & 69.2 & 75.1\\
    0 & 1 & 2 & 59.8 & 70.8 & 72.7\\
     \end{tabular}
        }
    \end{center}
    \label{tab:ablation:loss}
\end{table}

\begin{figure*}[!h]
  \centering
  \includegraphics[width=\linewidth]{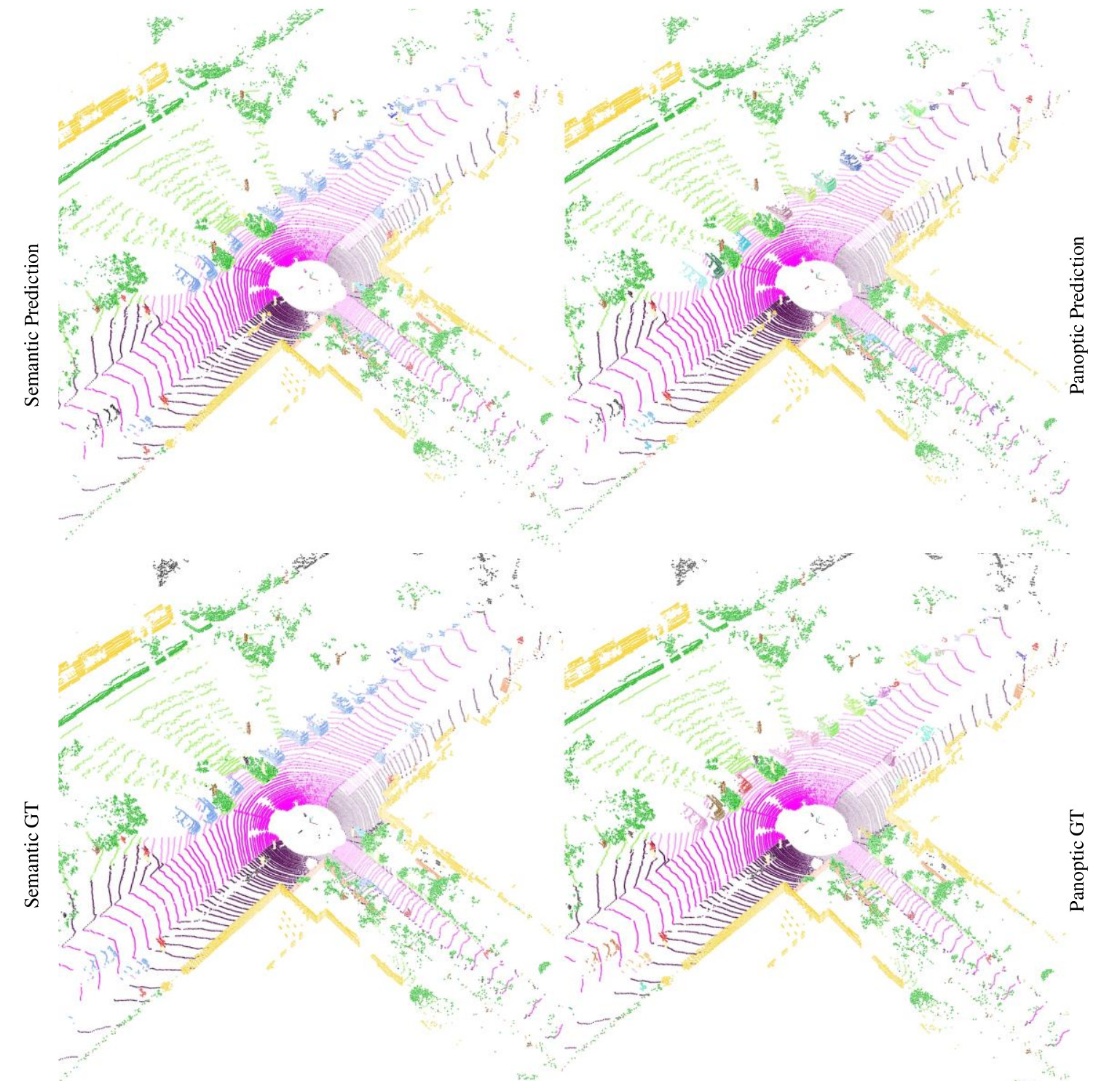}
  \caption{Qualitative examples in SemanticKITTI~\cite{semantickittipan}. Black in GT denotes the ignored class. Instances in panoptic results are randomly colored. Best viewed in color with zoom-in.}
  \label{fig:sem_pan}
\end{figure*}

\begin{figure*}[!h]
  \centering
  \includegraphics[width=\linewidth]{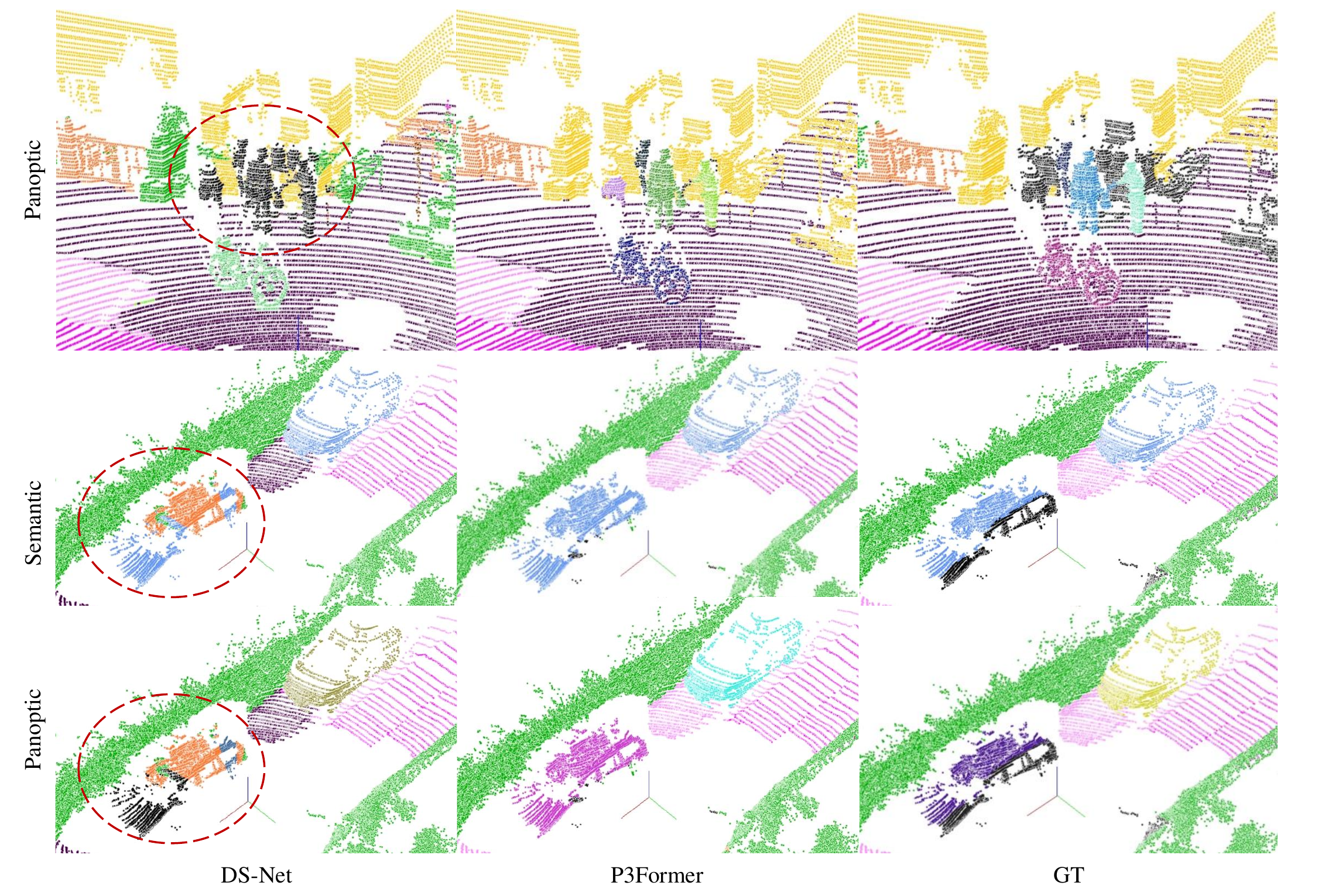}
  \caption{Qualitative comparisons in SemanticKITTI. Black in GT denotes the ignored class. Instances in panoptic results are randomly colored. Best viewed in color with zoom-in.}
  \label{fig:cmp}
\end{figure*}

\vspace{-1mm}

\section{Detailed Benchmarks}\label{sec:appendix:discussion}\vspace{-1mm}

We provide class-wise LiDAR panoptic results on SemanticKITTI~\cite{SemanticKITTI} test set (Table~\ref{tab:sem_test}), nuScenes~\cite{nuScenes} dataset (Table~\ref{tab:nus_val}).

\vspace{-1mm}

\section{Visualizations}\label{sec:appendix:discussion}
\begin{figure*}[!h]
  \centering
  \includegraphics[width=0.85\linewidth]{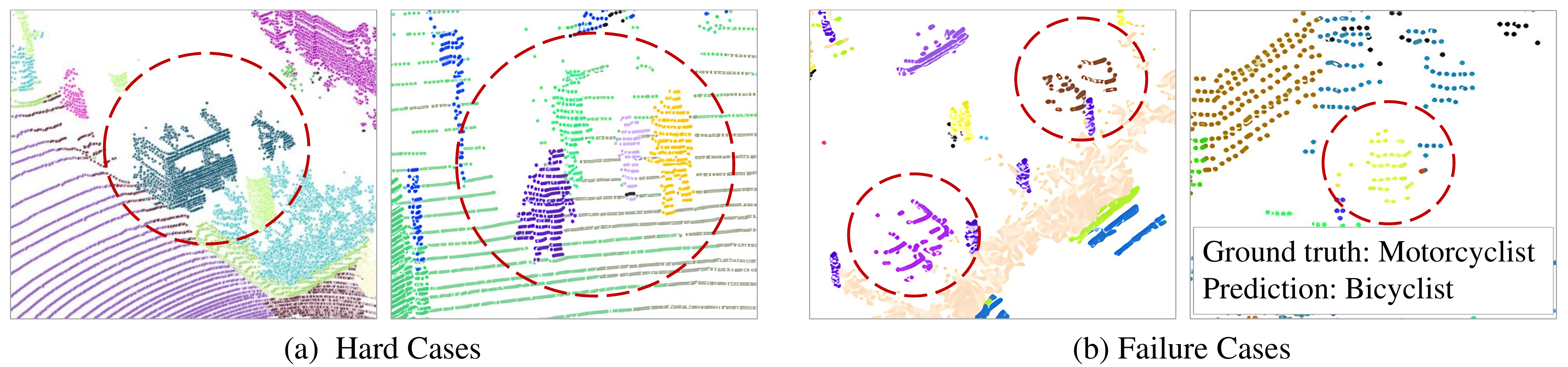}
  \caption{Hard Cases and Failure Cases. (a) P3Former successfully tackles the hard cases of truncated instances (left) and the dense crowd (right). (b) Failure cases of incomplete and close instances (left) and tail categories (right). Best viewed in color with zoom-in.}
  \label{fig:cases}
\end{figure*}

\vspace{-1mm}
\myparagraph{Qualitative Comparisons.}
We show the qualitative examples of semantic and panoptic segmentation on SemanticKITTI~\cite{SemanticKITTI} in Fig.~\ref{fig:sem_pan}. Our method performs well on large-scale outdoor scenes. Notably, P3Former can effectively handle the situations when instances are truncated because queries separate instances based on not only positional distances but also geometric similarity (Fig.~\ref{fig:cases}-(a, left)). Besides, our approach can tackle the crowded scenarios Fig.~\ref{fig:cases}-(a, right) since queries are specialized in precise positions.

We visualize the predictions of our method and other methods in Fig.~\ref{fig:cmp}.
Since recent SoTA~\cite{GPS3Net, SCAN, PHNet} methods haven’t released their code, we qualitatively compare our method to ~\cite{DS-Net} (CVPR 2021). The first row of Fig.~\ref{fig:cmp} shows that our methods are better at distinguishing close instances with similar geometry. The second and third rows of Fig.~\ref{fig:cmp} demonstrate that three-stage methods depend a lot on the results of semantic segmentation. If the semantic segmentation fails (the car circled in the second row), the panoptic segmentation is bound to fail. However, with a unified structure like P3Former, we can successfully eliminate this problem.

\myparagraph{Failure Cases.}
 However, P3Former still faces some challenges, such as failing to segment some incomplete and close instances (Fig.~\ref{fig:cases}-(b, left)), and wrongly classifying instances due to the highly imbalanced category distribution (Fig.~\ref{fig:cases}-(b, right)). We will investigate these inefficiencies in future studies.

\begin{figure*}[!h]
  \centering
  \includegraphics[width=0.90\linewidth]{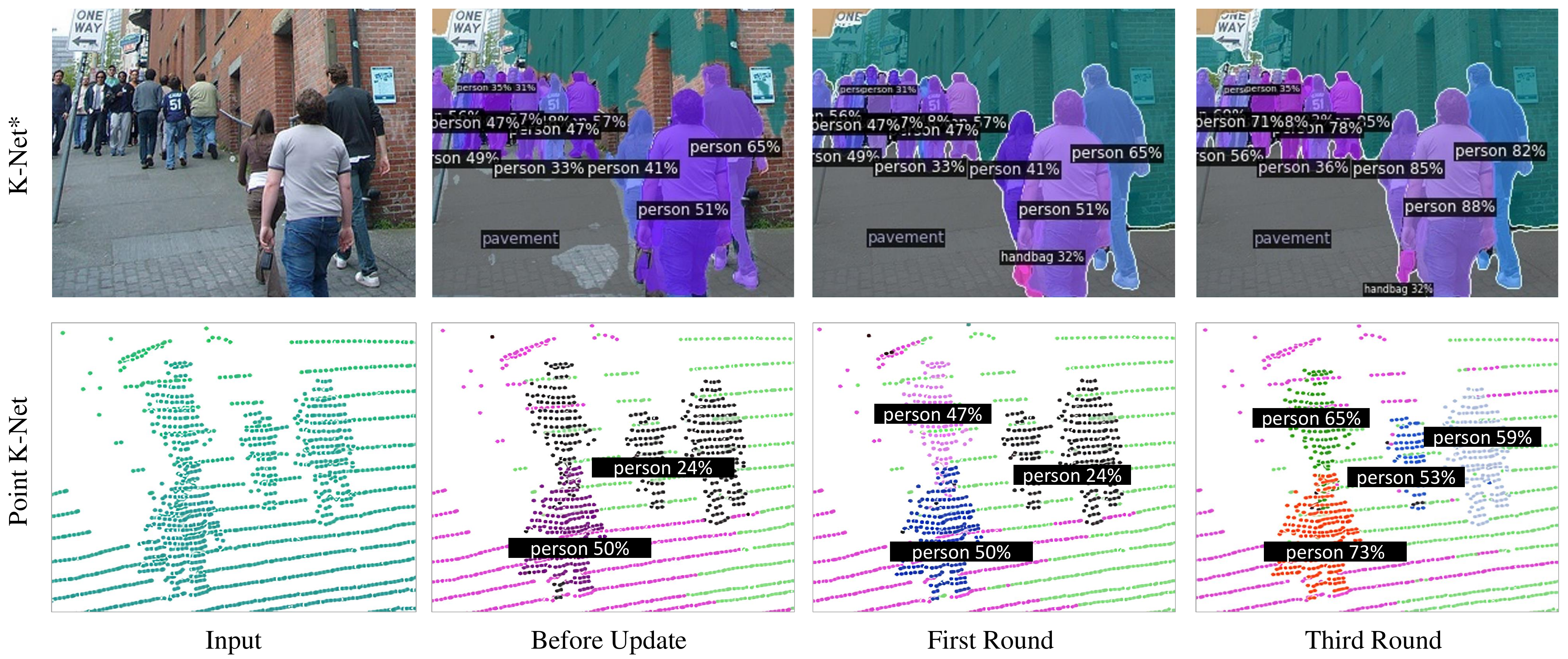}
  \caption{Comparisons of the update processes between K-Net and P3Former. Top line figures are borrowed from K-Net~\cite{K-Net}. Best viewed in color with zoom-in.}
  \label{fig:update}
\end{figure*}

\myparagraph{Visualization of Mask Update Processes.}
We compare the mask update processes between K-Net~\cite{K-Net} and P3Former in different domains (Fig.~\ref{fig:update}). It can be observed that, in images, instances have distinctive appearances, such as colors and textures. Each instance is expected to be assigned to one query before the update. Because instances are likely to overlap with each other, it is critical for queries to be adaptive to instances and refine their boundary during the update.

In contrast, instances in point clouds don't suffer from overlapping but are commonly geometrically alike. As a result, queries have difficulty separating instances with similar geometric structures and close positions before the update. With our mixed-parameterized positional encoding strategy and position-aware segmentation accentuating positional differences of instances, queries will gradually specialize on specific positions and successfully segment these instances.

\begin{table*}[t]
      \begin{center}  \caption{Ablation studies on Model Setting. Here $N_L$ indicates iterative layer numbers, $N_Q$ indicates the query number and $N_D$ indicates the feature dimension. ``$^{\dag}$": We measure the latency with the official codebase released by the authors on our hardware for reference.}
    \begin{threeparttable}
  \scalebox{0.90}{\tablestyle{8pt}{1.0}
    \begin{tabular}{c|cccc|cccc|cc}
    Model & $\mathrm{N_{L}}$ & $\mathrm{N_{Q}}$ & $\mathrm{N_{D}}$ & Backbone Size & PQ & RQ & SQ & IoU & Model Size & FPS \\
    \hline
    \specialrule{0.05em}{3pt}{3pt}
      Cylinder3D & & - & & 1x & 56.4 & 67.1 & 76.5 & 63.5 & - & -\\
      DS-Net     & & - & & 1x & 57.7 & 68.0 & 77.6 & 63.5   & 221M & $2.1^{\dag}$ \\
    \specialrule{0.05em}{3pt}{3pt} 
      ~\methodname  & 3 & 128 & 128 & 0.5x & 61.4 & 71.3 & 75.9 & 65.5 & 77M  & 19.7\\
                   & 6 & 128 & 256 & 0.5x & 62.5 & 72.4 & 76.1 & 66.5 & 120M & 14.2\\
                   & 6 & 128 & 256 & 1x   & 62.5 & 72.3 & 76.2 & 66.4 & 279M & 12.5\\
                   & 6 & 128 & 256 & 1.5x & 62.8 & 72.5 & 76.4 & 66.6 & 559M & 11.6\\
    \end{tabular}}
    \end{threeparttable}
    \label{tab:setting}
     \end{center}
      
\end{table*}

\begin{table*}[t]
  \begin{center}
  \caption{Class-wise LiDAR panoptic segmentation results on SemanticKITTI test set}
  \label{tab:sem_test}
  \scalebox{0.85}{\tablestyle{8pt}{1.0}
    \begin{tabular}{c|p{0.25cm}<{\centering}
                        p{0.25cm}<{\centering}
                        p{0.25cm}<{\centering}
                        p{0.25cm}<{\centering}
                        p{0.25cm}<{\centering}
                        p{0.25cm}<{\centering}
                        p{0.25cm}<{\centering}
                        p{0.25cm}<{\centering}
                        p{0.25cm}<{\centering}
                        p{0.25cm}<{\centering}
                        p{0.25cm}<{\centering}
                        p{0.25cm}<{\centering}
                        p{0.25cm}<{\centering}
                        p{0.25cm}<{\centering}
                        p{0.25cm}<{\centering}
                        p{0.25cm}<{\centering}
                        p{0.25cm}<{\centering}
                        p{0.25cm}<{\centering}
                        p{0.4cm}<{\centering}|c}
    
    Metrics & \rotatebox{90}{Car} & \rotatebox{90}{Truck} & \rotatebox{90}{Bicycle} & \rotatebox{90}{Motorcycle} & \rotatebox{90}{Other Vehicle} & \rotatebox{90}{Person} & \rotatebox{90}{Bicyclist} & \rotatebox{90}{Motorcyclist} & \rotatebox{90}{Road} & \rotatebox{90}{Sidewalk} & \rotatebox{90}{Parking} & \rotatebox{90}{Other Ground} & \rotatebox{90}{Building} & \rotatebox{90}{Vegetation} & \rotatebox{90}{Trunk} & \rotatebox{90}{Terrain} & \rotatebox{90}{Fence} & \rotatebox{90}{Pole} & \rotatebox{90}{Traffic Sign} & Mean \\
    
    \hline
    \specialrule{0.05em}{3pt}{3pt}
    PQ      & 93.0 & 47.3 & 52.1 & 65.6 & 61.1 & 75.2 & 79.4 & 63.4 & 89.2 & 69.4 & 53.0 & 19.8 & 89.4 & 81.4 & 63.2 & 47.7 & 57.6 & 58.5 & 66.7 & 64.9\\
    RQ      & 98.5 & 50.3 & 68.2 & 73.3 & 65.6 & 84.6 & 86.1 & 66.1 & 96.9 & 85.6 & 68.6 & 26.9 & 95.0 & 96.1 & 82.5 & 63.5 & 74.1 & 77.0 & 83.3 & 75.9 \\
    SQ      & 94.4 & 94.0 & 76.4 & 89.5 & 93.2 & 89.0 & 92.2 & 95.9 & 92.0 & 81.1 & 77.3 & 73.6 & 94.1 & 84.7 & 76.6 & 75.0 & 77.7 & 75.9 & 80.1 & 84.9 \\
    mIoU     & 95.6 & 49.2 & 54.8 & 66.7 & 57.8 & 69.9 & 72.0 & 38.9 & 91.5 & 75.8 & 67.2 & 40.1 & 92.1 & 86.1 & 72.4 & 69.8 & 68.5 & 64.0 & 65.6 & 68.3 \\
    
    \end{tabular}}
  \end{center}
\end{table*}

\begin{table*}[!h]
  \begin{center}
  \caption{Class-wise LiDAR panoptic segmentation results on nuScenes dataset}
  \label{tab:nus_val}
  \scalebox{0.80}{\tablestyle{8pt}{1.0}
    \begin{tabular}{c|p{0.25cm}<{\centering}
                        p{0.25cm}<{\centering}
                        p{0.25cm}<{\centering}
                        p{0.25cm}<{\centering}
                        p{0.25cm}<{\centering}
                        p{0.25cm}<{\centering}
                        p{0.25cm}<{\centering}
                        p{0.25cm}<{\centering}
                        p{0.25cm}<{\centering}
                        p{0.25cm}<{\centering}
                        p{0.25cm}<{\centering}
                        p{0.25cm}<{\centering}
                        p{0.25cm}<{\centering}
                        p{0.25cm}<{\centering}
                        p{0.25cm}<{\centering}
                        p{0.4cm}<{\centering}|c}
    
    Metrics & \rotatebox{90}{Barrier} & \rotatebox{90}{Bicycle} & \rotatebox{90}{Bus} & \rotatebox{90}{Car} & \rotatebox{90}{Construction Vehicle} & \rotatebox{90}{Motorcycle} & \rotatebox{90}{Pedestrian} & \rotatebox{90}{Traffic Cone} & \rotatebox{90}{Trailer} & \rotatebox{90}{Truck} & \rotatebox{90}{Driveable Surface} & \rotatebox{90}{Other Flat} & \rotatebox{90}{Sidewalk} & \rotatebox{90}{Terrain} & \rotatebox{90}{Manmade} & \rotatebox{90}{Vegetation} & Mean \\
    \hline
    \specialrule{0.05em}{3pt}{3pt}
    PQ      & 65.0 & 68.9 & 77.1 & 94.1 & 61.3 & 85.2 & 93.0 & 91.5 & 60.2 & 73.0 & 96.2 & 59.6 & 69.3 & 57.5 & 86.9 & 82.9 & 75.9 \\
    RQ      & 77.3 & 79.3 & 80.3 & 97.5 & 67.5 & 91.5 & 97.9 & 97.0 & 67.5 & 77.6 & 99.9 & 69.3 & 85.7 & 73.5 & 98.3 & 95.9 & 84.7 \\
    SQ      & 84.1 & 86.9 & 96.0 & 96.6 & 90.8 & 93.1 & 95.0 & 94.3 & 89.3 & 94.2 & 96.3 & 86.0 & 80.8 & 78.2 & 88.4 & 86.5 & 89.8 \\
    mIoU    & 68.2 & 40.3 & 92.4 & 93.2 & 57.0 & 84.1 & 76.3 & 65.1 & 73.2 & 85.3 & 96.5 & 71.5 & 74.1 & 74.8 & 89.6 & 87.2 & 76.8 \\
    
    \end{tabular}}
  \end{center}
\end{table*}

\clearpage

{\small
  \bibliographystyle{ieee_fullname}
  \bibliography{mainbib}
}

\end{document}